\documentclass[10pt,twocolumn,letterpaper]{article}

\usepackage[pagenumbers]{cvpr}

\usepackage[T1]{fontenc}
\usepackage{multirow}

\definecolor{tablebest}{RGB}{255,0,0}
\definecolor{tablesecond}{RGB}{0,0,255}
\newcommand{\bestscore}[1]{\textcolor{tablebest}{#1}}
\newcommand{\secondscore}[1]{\textcolor{tablesecond}{#1}}

\usepackage{placeins}
\definecolor{cvprblue}{rgb}{0.21,0.49,0.74}
\usepackage[pagebackref,breaklinks,colorlinks,allcolors=cvprblue]{hyperref}

\def\confName{CVPR}
\def\confYear{2027}

\title{CALE: Color-Adaptive Low-Light Enhancement\\
via Learnable Intensity Coordinates}
\author{%
Yi Ai\textsuperscript{1},
Zheng Chen\textsuperscript{1},
Yuanhao Cai\textsuperscript{2},
Yulun Zhang\textsuperscript{1}\thanks{Corresponding author.},
Xiaokang Yang\textsuperscript{1}\\
\textsuperscript{1}Shanghai Jiao Tong University\\
\textsuperscript{2}Johns Hopkins University
}

\begin{document}
\maketitle
\begin{abstract}
Small brightness differences in dark regions are difficult to recover, and their spacing in the coordinates used for restoration depends on the intensity mapping. A globally learned Box--Cox exponent controls the full response, coupling different brightness intervals and applying the same mapping to every image. We propose Color-Adaptive Low-Light Enhancement (CALE), which combines an image-conditioned intensity representation with structure-conditioned rectified flow (\mbox{SR-RF}). The representation, Box--Cox Intensity Adaptation (BIA), refines a normalized Box--Cox transform using a monotone piecewise-linear curve. Compact statistics of the low-light observation determine the output widths of brightness intervals, allowing the response to vary across images. Positive, normalized slopes preserve intensity order and endpoints and give the intensity mapping an analytic inverse. Combined with polar chromaticity, BIA defines the coordinates for the \mbox{SR-RF} network. The mapping and restoration network are trained jointly so that the reconstruction objective shapes the intensity response. Experiments on paired benchmarks and unseen datasets show competitive restoration performance across diverse low-light scenes, with qualitative comparisons illustrating the recovery of scene colors, contrast, and detail. Coordinate ablations separate the effects of interval flexibility and input conditioning, while Full and Tiny models offer quality--computation trade-offs.
\end{abstract}

\section{Introduction}
Low-light image enhancement must recover exposure while retaining scene colors, structural boundaries, and faint texture. Brightening can reveal noise and color errors, while aggressive smoothing can remove detail along with those degradations. Retinex-based networks \citep{retinexnet2018,kind2019,uretinex2022}, Transformers guided by illumination or signal-to-noise ratio \citep{retinexformer2023,snraware2022}, and generative restoration \citep{llflow2022,diffll2023,gsad2023} address these difficulties through decomposition, conditioning, and image priors. Each approach also relies on a representation in which predictions are formed and errors are measured. Small differences in dark regions occupy narrow intervals of raw intensity, and changing the representation changes how these differences enter the learning problem. We therefore focus on the coordinates used to learn image restoration.

\begin{figure}[t]
\centering
\includegraphics[width=\columnwidth]{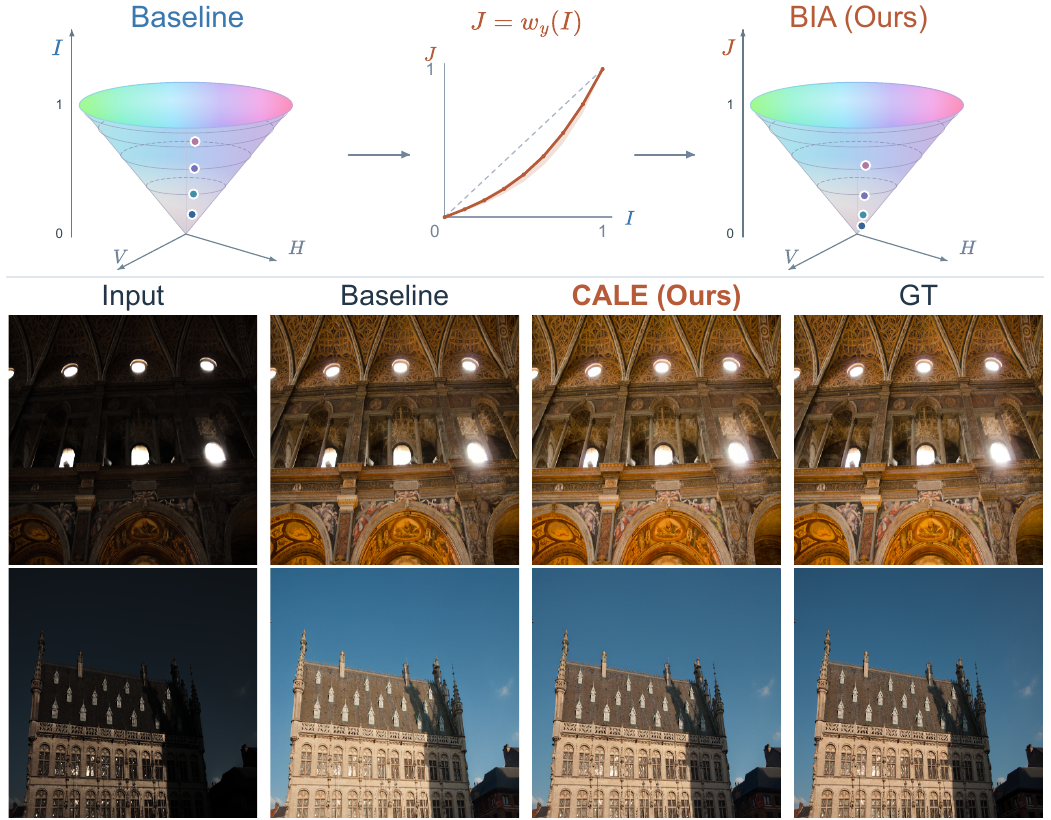}
\caption{Learned intensity coordinates for restoration. The schematic color-space views show how BIA refines the Box--Cox baseline with an image-conditioned monotone curve. Two Synthetic scenes show restoration with the same \mbox{SR-RF} network.}
\label{fig:firstpage-teaser}
\end{figure}

HVI provides a useful starting point by representing hue through polar chromaticity alongside an intensity axis \citep{hvi2025}. In the original-intensity reference used here, that axis is the maximum RGB channel. Consider two pixel pairs with the same small difference in this raw intensity, one in a shadow and the other in a highlight. They have equal separation along a linear intensity axis. A nonlinear response can give the shadow pair a larger separation, changing the numerical spacing of the restoration targets in that range. Its inverse determines how a prediction error in transformed intensity translates into decoded intensity. The forward and inverse responses connect coordinate design to restoration targets and local scalar decoding sensitivity.

The Box--Cox family offers a compact way to shape this response \citep{boxcox1964}. A globally learned exponent controls the full intensity range, coupling changes in different brightness intervals. Adjusting the exponent cannot allocate each interval's output width separately. Greater flexibility can be obtained by learning an allocation over several intervals, but a single shared allocation still applies to every image. A uniformly dark scene and a scene with bright lamps may distribute shadows, highlights, and useful detail differently, motivating a response that can adjust to the current image. We therefore study interval-level control and input conditioning as distinct design choices: the first changes the flexibility of the mapping, while the second allows its allocation to depend on the observed scene.

To provide this control, we introduce Box--Cox Intensity Adaptation (BIA), which composes a normalized Box--Cox transform with an image-conditioned monotone piecewise-linear curve. The global exponent sets the overall response, while compact statistics of the low-light observation predict the output widths of brightness intervals on the Box--Cox axis. Positive slopes and normalized widths preserve intensity order and endpoints and give the scalar intensity mapping an analytic inverse. Each image has one global curve, shared across spatial locations. Combining the refined intensity with polar chromaticity yields the restoration representation shown in Fig.~\ref{fig:firstpage-teaser}. The curve's role differs from direct enhancement: Zero-DCE applies predicted curves to produce an enhanced image \citep{zerodce2020}; BIA supplies coordinates for a separate restoration model.

Color-Adaptive Low-Light Enhancement (CALE) combines BIA with a structure-conditioned rectified flow (\mbox{SR-RF}) network. Using conditional flow matching \citep{rectifiedflow2023,flowmatching2023}, the network restores images in BIA coordinates with structural guidance from multiscale features of the low-light observation. The coordinate mapping and velocity network are optimized together, allowing the reconstruction objective to shape the intensity response for exposure recovery and perceptual fidelity. The observed input determines the transform used to encode training pairs and to decode the restored image at inference. A shared-curve control separates the benefit of learning brightness-interval allocation from the additional effect of input conditioning. We evaluate restoration quality on paired low-light benchmarks and unseen datasets, and compare the computational costs of the Full and Tiny models.

Our contributions to low-light restoration are:
\begin{itemize}
    \item We introduce BIA, an image-conditioned intensity mapping that controls brightness-interval widths and preserves monotonicity, with an analytic intensity inverse.

    \item CALE combines BIA with \mbox{SR-RF}, jointly learning the coordinates and network through flow matching and reconstruction.

    \item Experiments demonstrate competitive quantitative performance on paired benchmarks and unseen datasets, with qualitative comparisons showing faithful recovery of scene colors and local detail. Controlled coordinate ablations and comparisons of the Full and Tiny models assess the representation and its quality--computation trade-off.
\end{itemize}

\begin{figure*}[t]
\centering
\includegraphics[width=\textwidth]{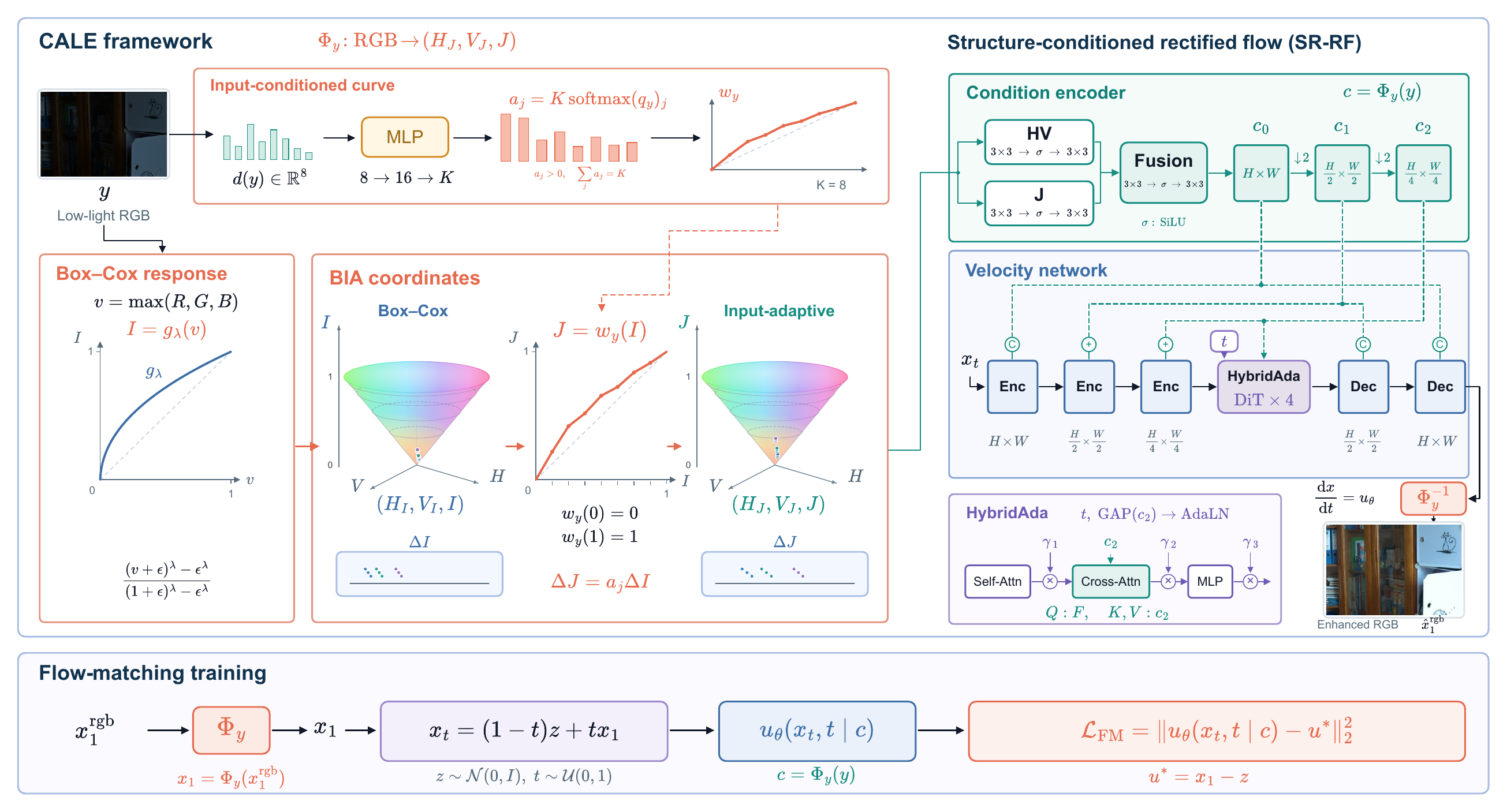}
\caption{Overview of CALE. BIA combines the Box--Cox transform with an image-conditioned curve to construct the restoration coordinates. In \mbox{SR-RF}, the velocity network receives condition features at matching spatial resolutions, and the inverse color transform decodes its endpoint to RGB. The training path and objective are shown below. Curves and points are schematic.}
\label{fig:architecture}
\end{figure*}

\section{Related Work}
\paragraph{Low-light restoration.}
Retinex-based methods decompose an observed image into reflectance and illumination, providing separate representations for scene appearance and exposure \citep{retinexnet2018,kind2019,kindplus2021}. Learned unfolding incorporates this decomposition into trainable iterative updates, combining an optimization structure with data-driven priors \citep{ruas2021,uretinex2022}. Other approaches relax the requirement for normally exposed training targets through no-reference losses, adversarial learning, or consistency between paired low-light observations \citep{zerodce2020,enlightengan2021,sci2022,pairlie2023,zeroig2024}. Transformers and state-space models extend the spatial context available for restoration \citep{llformer2023,retinexformer2023,snraware2022,mamballie2024,wavemamba2024}. Retinexformer, for example, uses illumination features to guide non-local interactions between differently lit regions. DarkIR and InterLight further study compound degradations and illumination priors \citep{darkir2025,interlight2026}. CALE focuses on the intensity mapping presented to the restoration network. BIA changes the spacing of brightness levels according to the observed image, while \mbox{SR-RF} models the spatial restoration process in those coordinates.

\paragraph{Generative restoration and flow matching.}
LLFlow models the conditional distribution of normally exposed images with a normalizing flow \citep{llflow2022}. Its invertible network maps training targets to a Gaussian distribution conditioned on the low-light observation; restoration follows the inverse direction. Diffusion models instead learn to reverse a gradual noising process \citep{ddpm2020}. Low-light adaptations introduce wavelet representations, Retinex decomposition, global structure, and generative perceptual priors \citep{diffll2023,diffretinex2023,gsad2023,lightendiffusion2024,gpp2025}. DiffLL combines conditional diffusion in a wavelet representation with high-frequency restoration. Diff-Retinex decomposes the input and uses generative diffusion to restore the illumination and reflectance components. Rectified flow and flow matching learn continuous velocity fields by regression along prescribed paths between source and target samples \citep{rectifiedflow2023,flowmatching2023}. This formulation separates the training objective from the numerical integration used at inference. CALE adopts flow matching in \mbox{SR-RF}, using a condition pyramid and a DiT-style bottleneck \citep{dit2023} to incorporate the observed scene.

\paragraph{Enhancement curves and color coordinates.}
Zero-DCE estimates image-specific, pixel-wise curves and applies them directly to the input to obtain an enhanced image \citep{zerodce2020}. Its curve design constrains the output range and monotonicity, and no-reference losses guide training without target images. Monotone splines also provide flexible invertible mappings in normalizing flows \citep{durkan2019spline}. Color-space design addresses a related choice: the distances and channel relationships seen by a restoration network. HVI uses polar chromaticity to address hue discontinuity and an intensity-dependent chromatic radius to reduce dark-region artifacts \citep{hvi2025,hviplus2025}. The Box--Cox family spans linear and logarithmic intensity laws \citep{boxcox1964} and has also been used in image preprocessing \citep{boxcoximage2020}. BIA retains polar chromaticity and combines a Box--Cox transform with an image-conditioned curve that assigns widths to brightness intervals. This curve defines restoration coordinates; its output is subsequently processed by \mbox{SR-RF} before RGB decoding. The shared-curve variant learns one allocation across images, allowing the role of input conditioning to be examined with the same curve family and predictor architecture.

\section{Method}
\subsection{Restoration in a learned coordinate system}
CALE couples the BIA color representation with the \mbox{SR-RF} network. Let $y\in[0,1]^{3\times H\times W}$ be a low-light image and $x_1^{\mathrm{rgb}}$ its normally exposed target. A color transform $\Phi_y$ defines
\begin{equation}
c=\Phi_y(y),\qquad x_1=\Phi_y(x_1^{\mathrm{rgb}}),\qquad
\hat x^{\mathrm{rgb}}=\Phi_y^{-1}(\hat x_1).
\label{eq:coordinate-consistency}
\end{equation}
The subscript identifies the observation that determines the color transform. Slopes predicted from $y$ are used to encode both $y$ and its normally exposed target; the inverse of this same transform decodes the restored endpoint. Thus the target is expressed in coordinates available from the low-light input at inference. The shared variant uses one color transform for all images. Figure~\ref{fig:architecture} shows how BIA and \mbox{SR-RF} form the complete restoration pipeline.

\subsection{Box--Cox intensity transform}
With $v=\max(R,G,B)$, normalized Box--Cox gives
\begin{equation}
I=g_\lambda(v)=
\frac{(v+\epsilon)^\lambda-\epsilon^\lambda}
     {(1+\epsilon)^\lambda-\epsilon^\lambda},
\quad \epsilon=10^{-3}.
\label{eq:boxcox}
\end{equation}
We learn $\lambda=0.05+0.95\sigma(\theta_\lambda)$. The transform preserves $0$ and $1$ and is strictly increasing. The family includes the linear response at $\lambda=1$ and has the logarithmic limit
\begin{equation}
\lim_{\lambda\to0}g_\lambda(v)=
\frac{\log(1+v/\epsilon)}{\log(1+1/\epsilon)}.
\label{eq:log-limit}
\end{equation}
Writing $D_\lambda=(1+\epsilon)^\lambda-\epsilon^\lambda$, the inverse is
\begin{equation}
g_\lambda^{-1}(I)=(I D_\lambda+\epsilon^\lambda)^{1/\lambda}-\epsilon.
\label{eq:inverse}
\end{equation}
For $\lambda<1$, $g_\lambda'(v)$ decreases with $v$. A small RGB intensity difference therefore spans a larger interval in a shadow than in a highlight. The exponent sets the strength of this effect across the entire raw-intensity range.

\subsection{Image-conditioned intensity mapping}

\label{sec:adaptive-coordinate}
A single Box--Cox exponent couples the response across the intensity range. We compose the transform with a piecewise-linear, image-conditioned curve $J=w_y(I)$ to control individual interval widths. For $K=8$ segments:
\begin{equation}
\begin{aligned}
a_j(y)&=K\,\operatorname{softmax}(q_y)_j,\\
w_y(I)&=a_1I+\sum_{j=1}^{K-1}(a_{j+1}-a_j)[I-j/K]_+.
\end{aligned}
\label{eq:adaptive-warp}
\end{equation}
Each input interval is $1/K$ wide and receives $a_j/K$ of the output axis. The positive slopes sum to $K$, so the allocated widths sum to one and the curve increases from $w(0)=0$ to $w(1)=1$. A slope of two, for example, gives that interval twice its Box--Cox width. Pixels within each brightness interval share a slope across the image.

\paragraph{Shared curve.}
A shared variant learns a single allocation across the dataset. An $8\!\rightarrow\!16\!\rightarrow\!8$ MLP with a SiLU activation takes a constant vector of ones and produces the logits $q$. Initializing its last layer to zero gives $a_j=1$ in every interval. Training thus starts at the Box--Cox baseline and learns the shared slopes jointly with the restoration network.

\paragraph{Input-adaptive curve.}
In the adaptive variant, the same MLP instead receives an eight-dimensional descriptor $d(y)$ computed from the observed RGB image: the three channel means, the mean and standard deviation of $v$, the proportions of pixels with $v<0.1$ and $v<0.02$, and mean HSV saturation. The descriptor covers the randomly cropped low-light patch during training and the full input image at inference. The output $q_y$ determines the interval widths for that image. The shared and adaptive predictors each use 280 parameters and the same curve parameterization, with constant or image-dependent inputs. At test time, the MLP weights stay fixed and the image descriptor determines the slopes.

\paragraph{Range allocation and inversion.}
Write the output width of interval $j$ and its upper boundary as
\begin{equation}
\beta_j(y)=\frac{a_j(y)}{K},\qquad
b_j(y)=\sum_{r=1}^{j}\beta_r(y),\qquad b_0=0.
\label{eq:range-allocation}
\end{equation}
The vector $\boldsymbol\beta$ lies in the probability simplex: its positive entries sum to one. It gives $K-1$ degrees of freedom for redistributing the unit output range. In raw intensity, the interval boundaries are $v_j=g_\lambda^{-1}(j/K)$. The intensity mapping is $T_y=w_y\circ g_\lambda$; for $v\in[v_{j-1},v_j]$ we have
\begin{equation}
T_y(v)=b_{j-1}+a_j\left(g_\lambda(v)-\frac{j-1}{K}\right).
\label{eq:composite-interval}
\end{equation}
Thus $T_y(v_j)=b_j$: the exponent sets the raw-intensity intervals, and the curve assigns their output widths. Pixels enter intervals according to their intensity; the image descriptor controls the allocation. Both stages preserve order and endpoints, making $T_y$ a continuous bijection of $[0,1]$. For $J\in[b_{j-1},b_j]$, inversion requires only
\begin{equation}
I=w_y^{-1}(J)=\frac{j-1}{K}+\frac{J-b_{j-1}}{a_j},\qquad
v=g_\lambda^{-1}(I).
\label{eq:adaptive-inverse}
\end{equation}

\paragraph{Polar chromaticity.}
With HSV hue $h$ and saturation $s$, we define the color representation as
\begin{equation}
\begin{aligned}
\Phi_y(R,G,B)&=(H_J,V_J,J),\\
(H_J,V_J)&=s\bigl(\sin(\pi J/2)+\delta\bigr)^k
                 \bigl(\cos(2\pi h),\sin(2\pi h)\bigr).
\end{aligned}
\label{eq:chromaticity}
\end{equation}
The shared scalar $k$ is learned from an initialization of $0.2$, without a positivity constraint; $\delta=10^{-8}$ is fixed. Refining $J$ changes the chromatic radius while retaining hue and saturation. The Box--Cox baseline sets $J=I$. For predicted endpoints, the decoder clamps $J$ to $[0,1]$ before Eq.~\eqref{eq:adaptive-inverse}, then clamps the recovered $I$ and $v$ to the same range. It also clips chromatic coordinates and saturation. The inverse uses detached values of $\lambda$ and $k$; curve slopes remain differentiable.

\subsection{Geometry of the learned coordinates}
\label{sec:coordinate-geometry}
Fix an observation and its intensity mapping. Within segment $j$, a small change in raw intensity gives
\begin{equation}
\frac{\mathrm{d}J}{\mathrm{d}v}=a_j g_\lambda'(v),\qquad
\frac{\mathrm{d}v}{\mathrm{d}J}
=\frac{D_\lambda}{\lambda a_j}(v+\epsilon)^{1-\lambda}.
\label{eq:inverse-derivative}
\end{equation}
The exponent controls the response across the intensity range, while each curve slope adjusts one brightness interval. A segment with $a_j>1$ expands intensity differences relative to the Box--Cox baseline, while its inverse reduces the corresponding scalar intensity sensitivity. This interval-wise adjustment is shared by all spatial locations in the image.

\subsection{Structure-conditioned rectified flow (SR-RF)}
With $x_1$ defined by Eq.~\eqref{eq:coordinate-consistency}, we sample $z\sim\mathcal{N}(0,I)$ and $t\sim\mathcal{U}[0,1]$ and use the standard flow-matching path \citep{rectifiedflow2023,flowmatching2023}:
\begin{equation}
x_t=(1-t)z+t x_1,\qquad u^\star=x_1-z.
\label{eq:rf-path}
\end{equation}
The velocity network minimizes
\begin{equation}
\mathcal L_{\mathrm{fm}}=
\mathbb E\,\operatorname{mean}\!\left[(u_\theta(x_t,t\mid c)-u^\star)^2\right].
\label{eq:flow-loss}
\end{equation}
\paragraph{Spatial condition pyramid.}
We encode the observed image's chromatic coordinates $(H_J,V_J)$ and intensity $J$ separately, then fuse their features into a condition pyramid:
\begin{equation}
\begin{aligned}
c_0&=F_c\bigl([E_{\mathrm{HV}}(H_J,V_J),E_J(J)]\bigr),\\
c_{j+1}&=D^c_j(c_j),\qquad j\in\{0,1\}.
\end{aligned}
\label{eq:condition-pyramid}
\end{equation}
Here $[\cdot,\cdot]$ denotes channel concatenation, and each $D^c_j$ halves the spatial resolution while doubling the channel width. The velocity encoder fuses $c_0$ with features of the evolving state $x_t$ at its input and adds $c_1,c_2$ at the corresponding coarser stages. During decoding, each upsampled feature is concatenated with both its encoder skip and the condition feature at the same scale. This supplies observed image structure at each resolution while the state features change with time. The velocity branch contains three encoder residual blocks and two decoder residual blocks; projected time embeddings enter every block.

\paragraph{Spatial and global conditioning.}
At the bottleneck, patch embeddings produce state tokens $F$ and condition tokens $C$ from the deepest encoder feature and $c_2$. Both receive two-dimensional sinusoidal positional embeddings. Cross-attention \citep{attention2017} uses state tokens as queries and condition tokens as keys and values, retaining the locations of observed features. The global vector $g_t=f_t(t)+W_p\operatorname{GAP}(c_2)$ combines the timestep with the pooled image condition. Linear maps in this notation include their learned biases. In each HybridAda block, a small predictor maps $g_t$ to three sets of channel-wise shifts, scales, and residual gates, $\{\eta_r,\alpha_r,\gamma_r\}_{r=1}^3$. Writing $\mathcal M_r(F)=(1+\alpha_r)\odot\operatorname{LN}_r(F)+\eta_r$, the block updates its state as
\begin{equation}
\begin{aligned}
F^{a}&=F+\gamma_1\odot\operatorname{SA}(\mathcal M_1(F)),\\
F^{b}&=F^{a}+\gamma_2\odot
\operatorname{CA}(\mathcal M_2(F^{a}),C),\\
F^{+}&=F^{b}+\gamma_3\odot\operatorname{FFN}(\mathcal M_3(F^{b})).
\end{aligned}
\label{eq:hybrid-update}
\end{equation}
Each block independently predicts modulation vectors broadcast over token positions. Their final prediction layer is initialized to zero, so the residual branches start with zero contribution. Bottleneck output tokens are mapped to a feature map and added to the deepest encoder feature.

\subsection{Training and inference}
We form an endpoint estimate from the single velocity prediction, $\hat x_1=x_t+(1-t)u_\theta(x_t,t\mid c)$, and supervise it in the transformed color coordinates and after RGB decoding:
\begin{equation}
\mathcal L=\mathcal L_{\mathrm{fm}}+
\omega_{\mathrm{rec}}\bigl(
\omega_{\mathrm{hvi}}\ell(\hat x_1,x_1)+
\ell(\Phi_y^{-1}(\hat x_1),x_1^{\mathrm{rgb}})\bigr).
\label{eq:total-loss}
\end{equation}
We set $\omega_{\mathrm{rec}}=0.5$ and $\omega_{\mathrm{hvi}}=0.1$, with
\begin{equation}
\ell=\alpha_1\ell_1+0.5\ell_{\mathrm{SSIM}}+50\ell_{\mathrm{edge}}+\alpha_P\ell_{\mathrm{VGG}}.
\label{eq:reconstruction-loss}
\end{equation}
The L1, SSIM \citep{ssim2004}, and Laplacian-residual MSE terms act directly on all three channels in each domain. VGG19 \citep{perceptualloss2016} receives the same affine preprocessing $A(q)=((q+1)/2-\mu)/\sigma$ in both domains, with ImageNet channel statistics. Coordinate channels are neither decoded to RGB nor clipped before these losses. We use $(\alpha_1,\alpha_P)=(1,1)$ on LOL-v1 and Synthetic, and $(2,0.3)$ on Real.

Training jointly updates the transform and network with one sampled time and one velocity evaluation. The transformed target $x_1$, path state $x_t$, and target velocity $u^\star$ retain gradients to the transform; the VGG target branch is detached.

At inference, we construct the color transform once from the observed image and keep it fixed during integration. Starting from $x^{(0)}=z$, Euler integration gives
\begin{equation}
\begin{aligned}
x^{(n+1)}&=x^{(n)}+\frac{1}{N}
u_\theta(x^{(n)},n/N\mid c),\\
\hat x^{\mathrm{rgb}}&=\Phi_y^{-1}(x^{(N)}).
\end{aligned}
\label{eq:coordinate-euler}
\end{equation}
The step count $N$ can be adjusted without retraining. We cache the condition pyramid, pooled condition, and cross-attention keys and values, while updating state queries and time-dependent modulation at each step.

An optional Tiny backbone retains condition width 64 and narrows the velocity branch to width 48 with 192-dimensional DiT blocks, using $1\times1$ projections to match feature widths. Tiny uses the same color transform and random-time objective as Full.

\section{Experiments}
\begin{figure*}[t]
\centering
\includegraphics[width=\textwidth]{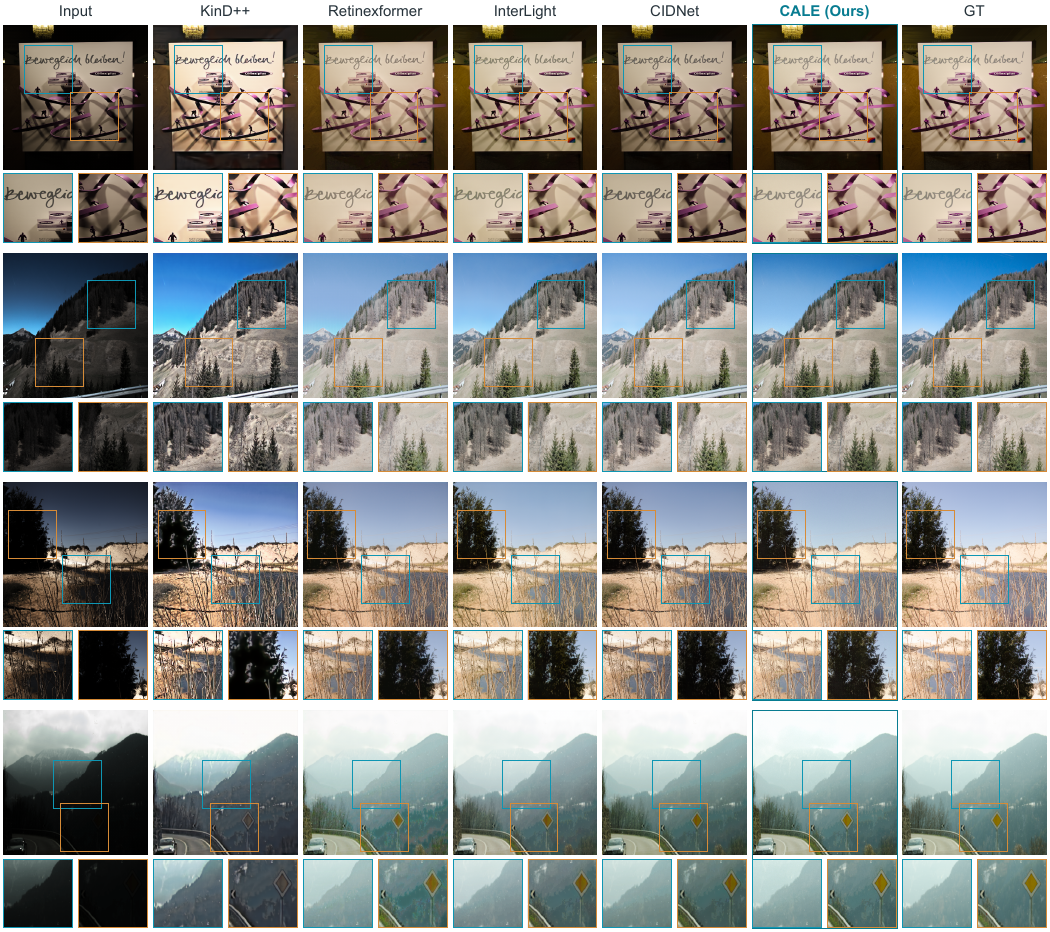}
\caption{Qualitative comparison on LOL-v2 Synthetic. CALE restores scene colors and local contrast. Colored boxes mark corresponding regions enlarged below each image for detailed comparison across methods.}
\label{fig:qualitative-main}
\end{figure*}

\begin{figure*}[t]
\centering
\includegraphics[width=\textwidth]{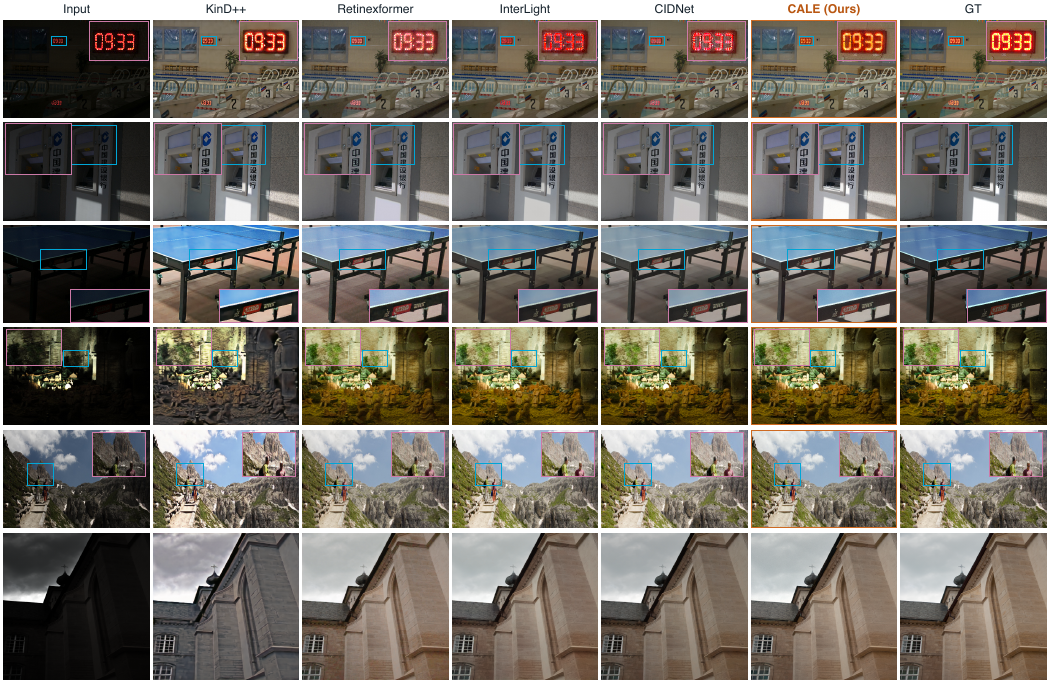}
\caption{Full images and aligned close-ups compare color, exposure, and local detail across low-light enhancement methods.}
\label{fig:qualitative-current}
\end{figure*}

\subsection{Evaluation protocol}
\label{sec:evaluation-protocol}
We evaluate on LOL-v1 and the Real and Synthetic splits of LOL-v2 \citep{retinexnet2018,lolv2_2021}, and assess cross-dataset generalization on DICM and LIME. PSNR, SSIM, and LPIPS \citep{ssim2004,lpips2018} are computed at native resolution against the corresponding reference images. We also report CLIP-IQA \citep{clipiqa2023} on the raw RGB outputs. In Table~\ref{tab:paired-benchmark}, CALE uses four Euler steps. LOL-v1 full-reference scores use GT-mean alignment, while LOL-v2 uses raw outputs.

LOL-v1 has 485 training and 15 test pairs. The Real and Synthetic splits of LOL-v2 have 689/100 and 900/100 pairs, respectively. We train the coordinate models from scratch for 1500 epochs with Adam \citep{adam2015} on $256\times256$ crops. Matched comparisons use the Full backbone, the same training recipe, FP32 and disabled TF32.

We report the step count for each experiment. Coordinate ablations use the same $N=4$ across variants, while the quality--efficiency comparison uses $N=1$ without endpoint fine-tuning. PSNR, SSIM and LPIPS for the coordinate study average three inference-noise realizations. No-reference image quality metrics are computed on the raw RGB predictions.

\subsection{Restoration quality and generalization}

\begin{table*}[t]
\caption{Paired LOL benchmark results. LOL-v1 PSNR, SSIM and LPIPS use GT-mean alignment; LOL-v2 scores and CLIP-IQA use raw predictions. The best and second-best scores are marked in \textcolor{tablebest}{red} and \textcolor{tablesecond}{blue}, respectively. Efficiency is compared in Table~\ref{tab:quality-efficiency}.}
\label{tab:paired-benchmark}
\centering
\scriptsize
\setlength{\tabcolsep}{4.1pt}
\resizebox{\textwidth}{!}{%
\begin{tabular}{lc*{12}{c}}
\toprule
\multicolumn{1}{c}{\raisebox{-2.25ex}[0pt][0pt]{Method}}
& \raisebox{-2.25ex}[0pt][0pt]{Params (M)}
& \multicolumn{4}{c}{LOL-v1}
& \multicolumn{4}{c}{LOL-v2 Synthetic}
& \multicolumn{4}{c}{LOL-v2 Real} \\
\cmidrule(lr){3-6}\cmidrule(lr){7-10}\cmidrule(lr){11-14}
& & PSNR$\uparrow$ & SSIM$\uparrow$ & LPIPS$\downarrow$ & C-IQA$\uparrow$
& PSNR$\uparrow$ & SSIM$\uparrow$ & LPIPS$\downarrow$ & C-IQA$\uparrow$
& PSNR$\uparrow$ & SSIM$\uparrow$ & LPIPS$\downarrow$ & C-IQA$\uparrow$ \\
\midrule
RetinexNet \citep{retinexnet2018} & 0.44 & 18.92 & 0.427 & 0.470 & 0.493 & 17.14 & 0.761 & 0.255 & \secondscore{0.571} & 16.10 & 0.401 & 0.543 & 0.542 \\
KinD \citep{kind2019} & 8.02 & 23.01 & 0.843 & 0.156 & 0.370 & 18.32 & 0.795 & 0.252 & 0.376 & 23.77 & \secondscore{0.874} & 0.121 & 0.400 \\
Zero-DCE \citep{zerodce2020} & 0.08 & 21.06 & 0.535 & 0.353 & 0.611 & 17.76 & 0.816 & 0.168 & 0.517 & 18.06 & 0.574 & 0.312 & \secondscore{0.578} \\
EnlightenGAN \citep{enlightengan2021} & 8.64 & 21.24 & 0.643 & 0.331 & 0.430 & 16.57 & 0.775 & 0.212 & 0.488 & 18.64 & 0.675 & 0.309 & 0.426 \\
KinD++ \citep{kindplus2021} & 8.27 & 24.16 & 0.842 & 0.155 & 0.409 & 19.26 & 0.806 & 0.226 & 0.447 & 22.21 & 0.843 & 0.175 & 0.397 \\
LLFlow \citep{llflow2022} & 38.86 & 25.18 & 0.873 & 0.115 & 0.397 & 23.43 & 0.934 & 0.050 & 0.499 & 17.43 & 0.832 & 0.176 & 0.331 \\
LLFormer \citep{llformer2023} & 24.55 & 26.11 & 0.830 & 0.165 & 0.331 & 17.16 & 0.787 & 0.244 & 0.410 & \bestscore{27.75} & 0.861 & 0.143 & 0.343 \\
Diff-Retinex \citep{diffretinex2023} & 58.91 & 27.06 & \bestscore{0.876} & 0.090 & 0.496 & 22.50 & 0.898 & 0.089 & 0.529 & 21.53 & 0.837 & 0.161 & 0.393 \\
Retinexformer \citep{retinexformer2023} & 1.61 & 27.17 & 0.850 & 0.129 & 0.427 & 25.67 & 0.930 & 0.059 & 0.511 & 22.79 & 0.840 & 0.171 & 0.324 \\
GSAD \citep{gsad2023} & 17.43 & 27.53 & \secondscore{0.875} & 0.092 & \secondscore{0.631} & 24.13 & 0.927 & 0.053 & 0.532 & 20.11 & 0.845 & \secondscore{0.113} & \bestscore{0.587} \\
LightenDiffusion \citep{lightendiffusion2024} & 27.83 & 23.75 & 0.830 & 0.175 & 0.294 & 21.61 & 0.866 & 0.155 & 0.405 & 22.96 & 0.855 & 0.166 & 0.278 \\
MambaLLIE \citep{mamballie2024} & 4.39 & -- & -- & -- & -- & \secondscore{25.87} & \secondscore{0.940} & 0.049 & 0.497 & 22.95 & 0.847 & 0.169 & 0.346 \\
DarkIR \citep{darkir2025} & 3.32 & 27.65 & \secondscore{0.875} & 0.115 & 0.361 & 25.55 & 0.935 & 0.052 & 0.490 & 23.88 & \bestscore{0.879} & 0.125 & 0.343 \\
CIDNet \citep{hvi2025} & 1.98 & 27.71 & \bestscore{0.876} & \secondscore{0.079} & 0.488 & 25.13 & 0.939 & \secondscore{0.045} & 0.520 & 23.90 & 0.866 & 0.122 & 0.503 \\
InterLight \citep{interlight2026} & 79.96 & \secondscore{27.87} & 0.873 & 0.093 & 0.485 & 25.80 & 0.935 & 0.050 & 0.510 & \secondscore{24.06} & 0.866 & 0.136 & 0.381 \\
\midrule
CALE (Ours) & 13.77 & \bestscore{27.98} & 0.868 & \bestscore{0.075} & \bestscore{0.648} & \bestscore{26.45} & \bestscore{0.941} & \bestscore{0.042} & \bestscore{0.576} & 23.87 & 0.864 & \bestscore{0.096} & 0.576 \\
\bottomrule
\end{tabular}
}
\end{table*}
On Synthetic, CALE reaches 26.45 dB in four steps (Table~\ref{tab:paired-benchmark}), compared with 25.13 dB for CIDNet, 25.67 dB for Retinexformer and 25.80 dB for InterLight. On Real, CALE obtains 23.87 dB, SSIM 0.864 and LPIPS 0.096. Its Real PSNR and SSIM do not lead this comparison. Across all three paired datasets, CALE achieves the lowest LPIPS among the compared methods, indicating close similarity to the reference images in the learned perceptual feature space. CALE also gives the highest CLIP-IQA on LOL-v1 and Synthetic.

Figure~\ref{fig:qualitative-main} shows how CALE recovers the poster's paper tone and purple ribbons, maintains contrast on the wooded hillside, and restores sky and water colors close to the reference. The mountain-road scene further illustrates exposure recovery under overcast light. Figure~\ref{fig:qualitative-current} extends the comparison to more severely underexposed indoor and outdoor inputs across the three datasets. These visual differences complement the LPIPS results, particularly in color fidelity and structural detail.

\begin{table*}[t]
\caption{Cross-dataset results on DICM (69 images) and LIME (10 images). CALE uses its LOL-v2 Real model with 20 Euler steps and no target-dataset training. All methods use the same inputs, proportionally resized to a maximum side of 768 pixels. BRISQ. and C-IQA denote BRISQUE and CLIP-IQA. Red and blue mark the best and second-best scores.}
\label{tab:unpaired-benchmark}
\centering
\scriptsize
\setlength{\tabcolsep}{5.0pt}
\renewcommand{\arraystretch}{1.03}
\begin{tabular}{llccccccc}
\toprule
Dataset & Method & NIQE$\downarrow$ & BRISQ.$\downarrow$ & MUSIQ$\uparrow$ & C-IQA$\uparrow$ & MANIQA$\uparrow$ & TOPIQ-NR$\uparrow$ & LOE$\downarrow$ \\
\midrule
\multirow{3}{*}{DICM} & Retinexformer \citep{retinexformer2023} & 4.19 & \bestscore{15.99} & 56.59 & 0.4245 & 0.2922 & 0.4230 & 303.72 \\
 & DarkIR \citep{darkir2025} & \bestscore{3.83} & 17.67 & \secondscore{60.73} & \secondscore{0.5439} & \bestscore{0.3811} & \secondscore{0.5074} & \bestscore{78.37} \\
 & CALE (Ours) & \secondscore{4.11} & \secondscore{17.14} & \bestscore{62.75} & \bestscore{0.5590} & \secondscore{0.3690} & \bestscore{0.5245} & \secondscore{94.40} \\
\midrule
\multirow{3}{*}{LIME} & Retinexformer \citep{retinexformer2023} & \bestscore{4.01} & \bestscore{8.52} & 61.92 & 0.4003 & 0.3597 & 0.4912 & \secondscore{97.80} \\
 & DarkIR \citep{darkir2025} & \secondscore{4.04} & 18.19 & \secondscore{62.87} & \secondscore{0.4494} & \secondscore{0.4213} & \secondscore{0.5320} & 99.03 \\
 & CALE (Ours) & 4.35 & \secondscore{15.67} & \bestscore{65.53} & \bestscore{0.5295} & \bestscore{0.4425} & \bestscore{0.5839} & \bestscore{93.93} \\
\bottomrule
\end{tabular}
\end{table*}

\paragraph{Generalization to unseen datasets.}
We apply the model trained on LOL-v2 Real directly to DICM and LIME. Table~\ref{tab:unpaired-benchmark} reports seven no-reference measures \citep{niqe2013,brisque2012,musiq2021,clipiqa2023,maniqa2022,topiq2024,loe2013}. CALE achieves the highest MUSIQ, CLIP-IQA and TOPIQ-NR on both datasets among the compared methods. It also obtains the highest MANIQA and lowest lightness-order error (LOE) on LIME. Agreement among these learned quality measures supports the perceptual benefit of CALE beyond the paired benchmarks, without training on either target dataset.

\begin{table}[!t]
\caption{Intensity coordinates on Synthetic with Full and four steps. Original intensity is the HVI reference with $I=v$; the controlled comparison uses the other three variants. Both curves add 280 parameters; no-reference metrics use raw RGB outputs.}
\label{tab:adaptive-study}
\centering
\footnotesize
\setlength{\tabcolsep}{2pt}
\renewcommand{\arraystretch}{1.0}
\setlength{\heavyrulewidth}{0.6pt}
\setlength{\lightrulewidth}{0.4pt}
\begin{tabular*}{\columnwidth}{@{\extracolsep{\fill}}lccccc@{}}
\toprule
Coordinate & PSNR$\uparrow$ & SSIM$\uparrow$ & LPIPS$\downarrow$ & BRISQ.$\downarrow$ & C-IQA$\uparrow$\\
\midrule
Original intensity & 25.58 & 0.9350 & 0.0451 & \secondscore{12.02} & \secondscore{0.5624}\\
Box--Cox & 25.98 & \secondscore{0.9388} & 0.0446 & 12.41 & 0.5560\\
Shared curve & \secondscore{26.07} & \bestscore{0.9409} & \bestscore{0.0410} & 12.80 & 0.5550\\
BIA & \bestscore{26.45} & \bestscore{0.9409} & \secondscore{0.0418} & \bestscore{11.63} & \bestscore{0.5765}\\
\bottomrule
\end{tabular*}
\end{table}

\begin{table}[!t]
\caption{SR-RF ablation on LOL-v2 Synthetic. Sep.: separate stems; MS: multiscale conditioning; CA: cross-attention; AdaLN: adaptive LayerNorm; ZI: zero-initialized residual gates.}
\label{tab:network-ablation}
\centering
\footnotesize
\setlength{\tabcolsep}{1.2pt}
\renewcommand{\arraystretch}{1.02}
\setlength{\heavyrulewidth}{0.6pt}
\setlength{\lightrulewidth}{0.4pt}
\begin{tabular*}{\columnwidth}{@{\extracolsep{\fill}}lcccccccc@{}}
\toprule
\multirow{2}{*}{Variant} & \multicolumn{5}{c}{Components} & \multicolumn{3}{c}{Restoration quality}\\
\cmidrule(lr){2-6}\cmidrule(lr){7-9}
& Sep. & MS & CA & AdaLN & ZI & PSNR$\uparrow$ & SSIM$\uparrow$ & LPIPS$\downarrow$\\
\midrule
w/o Sep. stems & -- & $\checkmark$ & $\checkmark$ & $\checkmark$ & $\checkmark$ & 23.92 & 0.925 & 0.060\\
w/o MS & $\checkmark$ & -- & $\checkmark$ & $\checkmark$ & $\checkmark$ & 24.63 & 0.930 & 0.053\\
w/o CA & $\checkmark$ & $\checkmark$ & -- & $\checkmark$ & $\checkmark$ & 25.11 & 0.936 & 0.047\\
w/o AdaLN & $\checkmark$ & $\checkmark$ & $\checkmark$ & -- & $\checkmark$ & 25.36 & 0.935 & 0.046\\
w/o ZI & $\checkmark$ & $\checkmark$ & $\checkmark$ & $\checkmark$ & -- & \secondscore{26.02} & \secondscore{0.938} & \bestscore{0.041}\\
\midrule
Full SR-RF & $\checkmark$ & $\checkmark$ & $\checkmark$ & $\checkmark$ & $\checkmark$ & \bestscore{26.45} & \bestscore{0.941} & \secondscore{0.042}\\
\bottomrule
\end{tabular*}
\end{table}

\begin{table}[!t]
\caption{One-step restoration quality and efficiency comparison.}
\label{tab:quality-efficiency}
\centering
\footnotesize
\setlength{\tabcolsep}{2pt}
\renewcommand{\arraystretch}{1.0}
\setlength{\heavyrulewidth}{0.6pt}
\setlength{\lightrulewidth}{0.4pt}
\begin{tabular*}{\columnwidth}{@{\extracolsep{\fill}}clccc@{}}
\toprule
Method & \multicolumn{1}{c}{PSNR / SSIM / LPIPS} & \begin{tabular}[c]{@{}c@{}}Params\\(M)\end{tabular} & GFLOPs & \begin{tabular}[c]{@{}c@{}}Latency\\(ms)\end{tabular}\\
\midrule
CALE-Full & \bestscore{26.00} / \bestscore{.9435} / \bestscore{.0377} & 13.77 & 168.34 & 35.51\\
CALE-Tiny & \secondscore{25.75} / \secondscore{.9404} / \secondscore{.0392} & 8.56 & 111.02 & 28.15\\
CIDNet \citep{hvi2025} & 25.13 / .9388 / .0450 & 1.98 & 16.36 & 24.44\\
\bottomrule
\end{tabular*}
\end{table}

\subsection{Component ablations}
The last three rows of Table~\ref{tab:adaptive-study} separate the effect of learning interval widths from that of conditioning them on the image. These three variants use the same \mbox{SR-RF} network and training recipe; the original-intensity HVI result is included as a separate reference. Learning eight shared interval widths improves PSNR, SSIM, and LPIPS over Box--Cox. Conditioning the same predictor on the image descriptor raises PSNR by a further 0.379 dB. BIA has the highest PSNR across all four coordinates, while the shared curve has the lowest LPIPS and ties BIA in SSIM at the reported precision. BIA also has the lowest BRISQUE and highest CLIP-IQA among these coordinates.

\paragraph{Restoration network.}
Table~\ref{tab:network-ablation} separates the contributions of the five \mbox{SR-RF} components by removing each from the full network in turn.

\subsection{Quality versus computation}
Table~\ref{tab:quality-efficiency} compares single-step CALE-Full and CALE-Tiny with CIDNet on LOL-v2 Synthetic. Both CALE variants use BIA and random-time flow-matching training without endpoint fine-tuning. Relative to Full, Tiny reduces the counted FLOPs from 168.34 G to 111.02 G and latency from 35.51 ms to 28.15 ms, while improving all three quality metrics over CIDNet at a similar measured latency (24.44 ms). Timing includes both color transforms, the condition encoder and solver on an A6000 at $256\times256$, batch size one, FP32, with TF32 disabled. We report the median of 90 synchronized measurements after ten warm-ups and count convolution, linear and attention matrix products at two operations per MAC.

\FloatBarrier
\section{Conclusion}
We presented CALE, a low-light enhancement framework that learns the intensity representation together with the \mbox{SR-RF} model. BIA extends the normalized Box--Cox response through image-conditioned allocation of brightness-interval widths, while retaining monotonicity and an analytic intensity inverse. Evaluations on paired benchmarks and unseen datasets show competitive restoration quality, with a perceptual similarity advantage on LOL-v2 Real as measured by LPIPS. Coordinate ablations distinguish the contribution of interval flexibility from that of input conditioning, and the Full and Tiny variants offer different quality--computation trade-offs.

\section*{LLM Usage Statement}
OpenAI GPT assisted with manuscript drafting, code implementation and debugging, visualization preparation, and LaTeX formatting.

\clearpage
{
    \small
    \bibliographystyle{ieeenat_fullname}
    \bibliography{main}
}

\end{document}